\documentclass[11pt]{article}

\usepackage[preprint]{acl}

\usepackage{times}
\usepackage{latexsym}
\usepackage{multirow}
\usepackage[T1]{fontenc}

\usepackage[utf8]{inputenc}

\usepackage{microtype}

\usepackage{inconsolata}

\usepackage{graphicx}
\usepackage{amsmath}
\usepackage{booktabs}
\usepackage{fancyvrb}
\usepackage{fvextra}
\DefineVerbatimEnvironment{agentbox}{Verbatim}{fontsize=\small,breaklines=true,breakanywhere=true}

\title{HERO: Human-profile Enhanced Retrieval Optimization Framework for Long-term Agent Memory}

\author{
  Yuanhua Lin \and Yile Li \and Zhiyuan Zhao \and Jing Shang \and Jian Sun \\
  China Mobile Information Technology Co., Ltd., Beijing, China \\
  \texttt{\{linyuanhua, liyile, zhaozhiyuan, shangjing, sunjian01\}@chinamobile.com}
}

\begin{document}
\maketitle
\begin{abstract}
Long-term memory is crucial for personalized responses and long-horizon agent interactions. Existing methods often rely on LLMs to compress or rewrite dialogue histories and use the transformed memories as retrieval evidence. Despite the progress in organizing fragmented
contexts, two major drawbacks persist: (1) information loss from compression, which discards fine-grained but later useful details, and (2) semantic drift from rewriting, which erodes the original tone and situated context. 
In this work, we propose a novel Human-profile Enhanced Retrieval Optimization framework for long-term agent memory (HERO). Specifically, HERO converts the dialogue history into a traceable heterogeneous memory graph that preserves raw dialogue text as evidence for reasoning, thereby mitigating information loss. For retrieval, HERO extracts initial anchors from the current query and incorporates human profiles via an iterative graph traversal; these anchors and profiles provide guidance signals that adaptively activate the most informative regions of the graph. Experiments on two benchmark datasets  show that HERO outperforms strong baselines on both factual and personalized reasoning, while providing more faithful access to raw dialogue evidence. 
\end{abstract}

\section{Introduction}

The improvement of Large Language Models (LLMs) has prompted the evolution of agents~\cite{Schlegelarticle, DBLP:conf/nips/YangJWLYNP24, DBLP:journals/frai/PantiukhinSKJK25}. Agent memory is a core capability for such systems, enabling agents to reuse experiences and sustain personalized long-term interactions~\cite{DBLP:journals/tois/ZhangDBMLCZDW25, DBLP:journals/corr/abs-2508-07407}. A common implementation is Retrieval-Augmented Generation (RAG)~\cite{DBLP:conf/nips/LewisPPPKGKLYR020, DBLP:BorgeaudMHCRM0L22}, which retrieves historical context or external knowledge as model input. 
While RAG helps mitigate hallucination~\cite{DBLP:journals/tois/HuangYMZFWCPFQL25}, it remains a memory mechanism that is decoupled from the reasoning process. The retrieved isolated text fragments lack inherent structured connections, leading to substantial redundancy and partially relevant content. Consequently, the standard RAG paradigm is akin to static knowledge access from an external knowledge base rather than a human-like memory system, and thus lacks the capacity for multi-turn, dynamic interaction and reasoning during the cognitive process.

To overcome the aforementioned bottlenecks, episodic long-term memory~\cite{articleMargaret2006, DBLP:conf/ro-man/KasapM10, DBLP:journals/corr/abs-2502-06975} has emerged as a promising research direction. It focuses on the dynamic integration of memory and reasoning, aiming to achieve flexible activation, adaptive reorganization, and continuous evolution of memory. 
From the perspective of memory organization, existing methods can be broadly divided into summary-based and graph-based methods. Summary-based methods refine refine dialogue histories to generate summaries or human profiles~\cite{LangChain,  DBLP:conf/emnlp/00110PCML0024, DBLP:conf/icml/LeeCFCF24}; graph-based methods extract memory elements (such as entities and keywords) or leverage LLMs to summarize text blocks, and represent temporal, causal, or semantic associations via relational edges~\cite{DBLP:conf/iclr/SarthiATKGM24, DBLP:conf/nips/GutierrezS0Y024, DBLP:journals/corr/abs-2404-16130,  DBLP:conf/iclr/RezazadehLWB25}. These methods address the fragmented retrieval and redundancy inherent in RAG. Nevertheless, both approaches diverge from human cognitive science~\cite{articleDickerson2009,articleTanguay2023}: the efficacy of agent memory is fundamentally constrained by the narrative and essential memory extraction capabilities of LLMs. For instance, existing graph-based methods are prone to introducing noise and ambiguity when constructing the topology due to inaccurate relation extraction ~\cite{DBLP:journals/corr/abs-2501-00309,DBLP:journals/corr/abs-2410-05779}, hierarchical clustering may be compromised by erroneous LLM-generated summaries, causing hallucinations to propagate from lower-level representations to higher-level abstractions\cite{DBLP:journals/corr/abs-2510-05520}.

This reliance on abstraction causes two closely related problems. First, compression may omit details that appear unimportant at write time but later become crucial, such as temporal references, causal conditions, numerical facts, or subtle preference changes. Second, rewriting dialogue into summaries, profiles, or standardized memory units may introduce semantic drift: even when the main fact is retained, the user's original wording, affect, speaker attribution, or situated context can be weakened. 

Based on the above analysis, a natural idea is to construct a long-term memory that inspired by human cognition for agents, which can swiftly locate relevant concepts (including basic attributes, preferences, etc.), activate related episodic memory slices, perform factually faithful reasoning and generate personalized responses. 
In this work, we propose \textbf{HERO}, a Human-profile Enhanced Retrieval Optimization framework for long-term agent memory. HERO follows a human-cognition-inspired principle: raw dialogue traces are preserved as episodic evidence, while higher-level profile information is used only to guide memory activation. Specifically, HERO constructs a heterogeneous memory graph with Episodic Traces, Episodic Units, and Episodic Cues, connected through structural inclusion, semantic indexing, and temporal chaining. This graph preserves local dialogue contexts while supporting traversal across fine-grained textual evidence and adjacent chronological contexts. In parallel, HERO maintains profile insights derived from episodic traces, but these profile nodes act as navigational bridges rather than replacement memories.

For retrieval, we introduce an iterative cue-activation mechanism to accommodate distinct memory tendencies for factual and personalized reasoning. Specifically, we operate across two dimensions—query relevance and implicit reasoning based on human-profiles, generating cues across three levels: queries, episodic units, and profiles. This extends the naive memory retrieval process, which relies solely on query-embedding similarity, into a two-stage pipeline. First, we leverage human profiles to guide cue reconstruction and expansion of the original query; then, we perform path-based retrieval on the memory graph using the generated cues. In this way, our model can retrieve explicit episodic memories and trace implicit, cross-episodic long-range dependencies through the graph structure. 

The contributions of this paper are as follows:
\begin{itemize}
\item We analyze the limitations of existing long-term memory approaches and propose a novel human-profile-enhanced retrieval optimization framework, which incorporates human-cognition-inspired design principles.
\item We design a heterogeneous memory graph that preserves complete dialogue episodes as final evidence and uses human-profile information as retrieval guidance rather than memory replacement.
\item We design a profile-aware cue activation and graph retrieval mechanism that retrieves explicit facts and implicit cross-episodic dependencies while constraining topic drift through query-conditioned filtering. 
\item We conduct experiments on two benchmark datasets. HERO achieves 56.06\% F1 on LoCoMo and 70.63\% accuracy on PERSONAMEM, demonstrating that HERO outperforms strong baselines on complex factual question answering and personalized reasoning tasks.
\end{itemize}

\section{Related Work}
To address the challenge of processing long-horizon interaction histories in agents, recent research on long-term memory has broadly diverged into two strands. 
One line is the summary-based approaches~\cite{LangChain, DBLP:journals/corr/abs-2506-06326, DBLP:journals/corr/abs-2308-08239, DBLP:conf/emnlp/00110PCML0024, DBLP:conf/icml/LeeCFCF24, DBLP:journals/corr/abs-2310-08560, journals/corr/abs-2510-18866, DBLP:journals/corr/abs-2505-22101, DBLP:conf/iclr/PanWJLCL0LZQ025, DBLP:conf/naacl/LiYZDWC25}. 
The main idea is to refine original texts into hierarchical summaries, or structured facts to bypass the context window limitation. 
Representative systems like MemoryBank~\cite{ZhongGGYW24}, LightMem~\cite{journals/corr/abs-2510-18866}, EverMemOS~\cite{2026arXiv260102163H}, Mem0~\cite{journals/corr/abs-2504-19413}, and CAM~\cite{DBLP:journals/corr/abs-2510-05520} implement various cognitive-inspired write-time compression or memory lifecycle management schemes. However, these methods inevitably suffer from information loss and semantic drift due to their heavy reliance on LLMs to rewrite or summarize raw interactions.

Another line is the graph-based memory approaches~\cite{DBLP:conf/iclr/SarthiATKGM24, DBLP:journals/corr/abs-2502-12110, DBLP:journals/corr/abs-2410-05779, DBLP:journals/corr/abs-2501-00309, DBLP:conf/ijcai/AnokhinSSEK0B25, DBLP:journals/corr/abs-2506-07398, DBLP:conf/nips/GutierrezS0Y024, DBLP:journals/corr/abs-2404-16130, DBLP:journals/corr/abs-2510-10114, DBLP:conf/iclr/RezazadehLWB25,DBLP:journals/corr/abs-2501-13956}, which extracts structured knowledge from text to construct graphs and performs path-based retrieval. This structure naturally supports multi-hop reasoning and alleviates knowledge conflicts through temporal modeling. 

These methods significantly advance long-term agent memory. However, they primarily rely on compressing historical information and the re-narration power of LLMs, which inevitably leads to irreversible information loss or semantic distortion. 
In this paper, we construct a novel heterogeneous memory graph, which is faithful to the original text. Moreover, we leverage the human-profile and the current query through a multi-round iterative to generate initial cues, which in turn guide agents' personalized response.

\begin{figure*}[t]
  \centering
  \includegraphics[width=\linewidth]{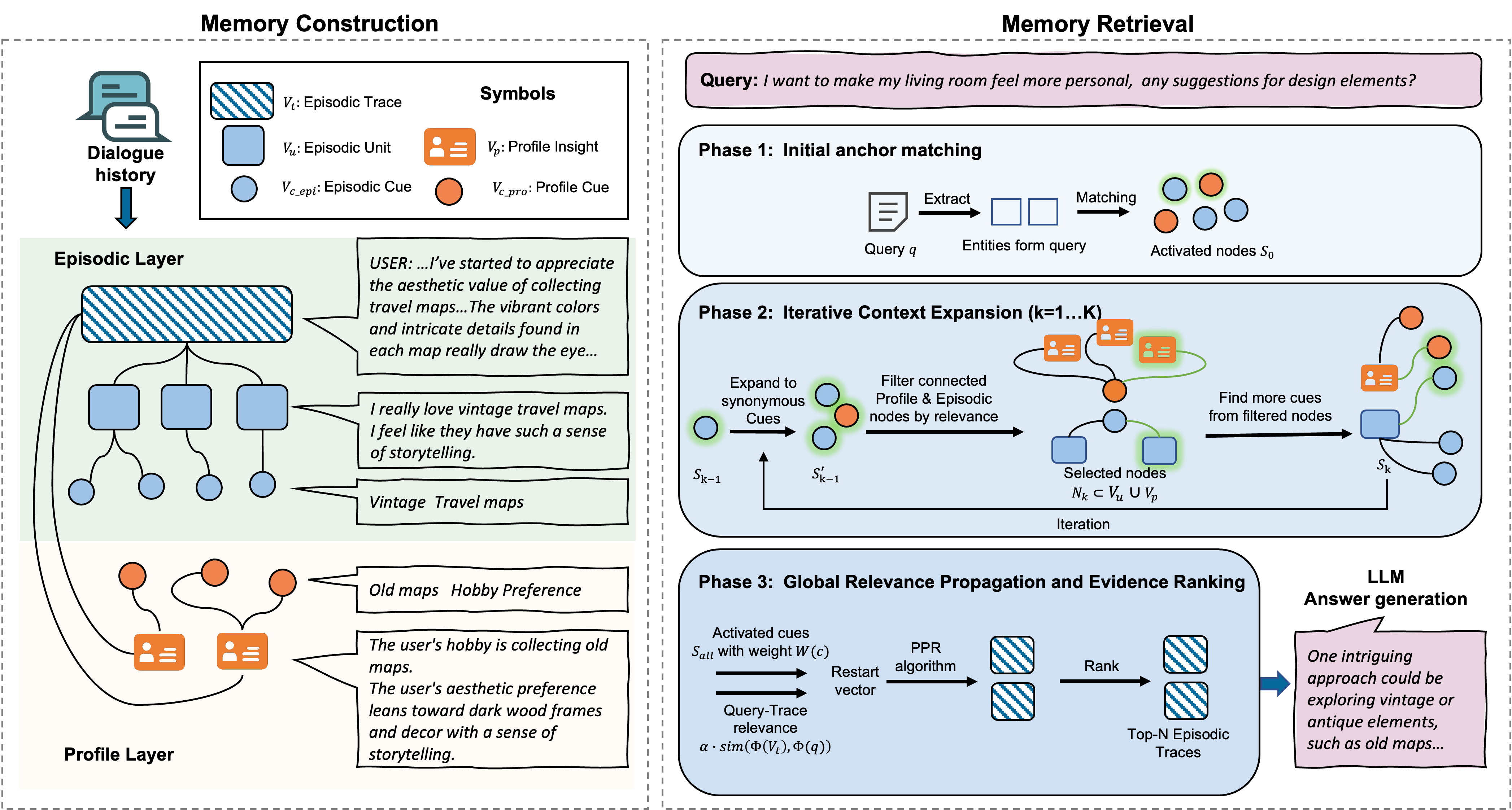}
    \caption{Overview of the proposed HERO framework. Left: Memory Construction. User dialogues are processed to extract Episodic Traces, which are further organized into Episodic Units and Episodic Cues to form the Episodic Layer. In parallel, Profile Insights and corresponding Profile Cues are derived from Episodic Traces to construct the Profile Layer. Episodic Cues and Profile Cues jointly constitute the cue nodes in a heterogeneous memory graph. Right: Memory Retrieval. Given a user query, entities are first extracted and matched against the cue nodes in the graph to obtain an initial candidate set. Through iterative expansion via Cue–Episodic Unit and Cue–Profile Insight bridges, additional relevant cue nodes are activated and incorporated into the Personalized PageRank initialization. In this process, Episodic Units support explicit semantic linking, while Profile Insights provide implicit, profile-consistent semantic connections. }
  \label{fig:hero}
\end{figure*}

\section{Methodology}

We define the user's dialogue sessions as $\mathcal{H}=\{s_1, s_2, \ldots, s_T\}$, where each session $s_i$ consists of a sequence of events or multi-turn dialogues. We assume the existence of a latent human profile $\mathcal{P}$, representing the user's accumulated facts, evolving traits, and preferences derived from $\mathcal{H}$.

Given a current user query $q$, the goal of HERO is to retrieve an optimal memory subset $\mathcal{M}^* \subset \mathcal{H}$ that maximizes the generation probability of the target response $r$. This optimization problem can be formulated as:
\begin{equation}
    \mathcal{M}^* = \underset{\mathcal{M} \subset \mathcal{H}}{\arg\max} \ P(r \mid q, \mathcal{M}, \mathcal{P}; \Phi_{\text{LLM}}),
\end{equation}
where $\Phi_{\text{LLM}}$ denotes the parameters of the backbone Large Language Model. 

We propose HERO, a long-term agent memory framework that enables personalized reasoning without sacrificing the fidelity of raw dialogue records. As shown in Figure~\ref{fig:hero}, HERO has two components: (1) Memory Construction, which organizes multi-granular conversational experiences into a coarse-to-fine heterogeneous graph; and (2) Memory Retrieval, which uses multi-level cues to retrieve profile-consistent evidence for explicit fact retrieval and implicit cross-session reasoning. 

\subsection{Memory Construction}

To support factually faithful reasoning over episodic memory, HERO constructs an episodic-centered heterogeneous graph $\mathcal{G} = (\mathcal{V}, \mathcal{E})$. The graph has two layers: an \textit{Episodic Layer} that stores raw historical events and a \textit{Profile Layer} that distills human profiles.

The Episodic Layer focuses on faithfully recording what happened in the original dialogues. It hierarchically segments the conversation flow into granular units to ensure memory traceability, and contains three types of nodes: Episodic Traces ($\mathcal{V}_t$), Episodic Units ($\mathcal{V}_u$), and Episodic Cues ($\mathcal{V}_{c\_epi} \subset \mathcal{V}_c$):

(1) Episodic Traces are partitioned into semantically coherent sessions or time windows. Each trace node $v_t \in \mathcal{V}_t$ represents a local conversation scenario memory, encompassing the full original text, timestamp, and sequential index. 

(2) Episodic Units are the further segmented atomic sentence-level units $v_u$. Each unit node $v_u \in \mathcal{V}_u$  acts as the fundamental carriers of evidence, retaining detailed contextual constraints. 

(3) Episodic Cues are entities and keywords that extracted from $v_u$ by LLMs, and serve as associative bridges. Besides, entities extracted from different units are normalized and merged into shared nodes in the global cue set $\mathcal{V}_c$.

The Profile layer is designed for understanding users. This layer performs  high-level semantic abstraction to model users' implicit attributes, and contains two types of node: Profile Insights ($\mathcal{V}_p$) and Profile Cues ($\mathcal{V}_{c\_pro} \subset \mathcal{V}_c$):

(1) Profile Insights ($\mathcal{V}_p$): For each episodic trace, the LLM distills human-centric information: specifically \textit{Facts}, \textit{Insights}, and \textit{Traits} to generate profile nodes $v_p$, where $v_p \in \mathcal{V}_p$ . 
    
(2) Profile Cues ($\mathcal{V}_{c\_pro} \subset \mathcal{V}_c$): Fine-grained entities are further extracted from $v_p$ to link abstract user portraits back to the global semantic space. Note that $\mathcal{V}_c = \mathcal{V}_{c\_epi} \cup \mathcal{V}_{c\_pro}$.

The final node set $\mathcal{V}$ is composed of multi-granular semantic abstractions: 
\begin{equation}
\mathcal{V} = \mathcal{V}_t \cup \mathcal{V}_u \cup \mathcal{V}_p \cup \mathcal{V}_c.
\end{equation}

The edge set $\mathcal{E}$ for topological construction contains three types:

(1) Structural Inclusion: Directed edges $\{(v_u, v_t), (v_p, v_t)\}$ link fine-grained units and profile nodes to their source episodic traces. Directed edges $\{(v_c, v_t), (v_c, v_p)\}$ connect cue nodes to their originating episodic traces and profile insights.

(2) Semantic Indexing: Weighted edges $\{(v_c, v_t), (v_c, v_p)\}$ connect retrieval cues with episodic traces and profile insights; weights $w_{c,t}$ and $w_{c,p}$ are based on occurrence frequency.

(3) Temporal Chaining: Adjacent episodic traces are connected by $\{(v_{t_i}, v_{t_{i+1}})\}$ to preserve temporal order.

This construction turns streaming conversations into a traversable cognitive association graph with three benefits. First, the Episodic Layer supports faithful retrieval of raw evidence and multi-hop associations across sessions, enabling variable-resolution retrieval from coarse traces to fine-grained facts without irreversible compression. Second, the Profile Layer captures user preferences and provides implicit links for retrieving profile-consistent evidence even without direct keyword overlap. Third, HERO supports incremental updates: new dialogues create local nodes and edges (linking to the latest predecessor and merging new cues), without full graph reconstruction.

\subsection{Memory Retrieval}

During retrieval, HERO employs a profile-enhanced retrieval optimization strategy to handle two query types. Factual questions activate query-related cues to locate complete raw evidence and avoid losses from extraction or summarization. Personalized reasoning questions activate profile-related cues, because their target attributes may not appear as explicit entities in the dialogue. For example, answering "Describe the user's dietary preferences" requires profile insights to bridge abstract preferences with multiple raw fragments.
HERO retrieves evidence in three phases: (1) Initial Anchor Identification, (2) Iterative Context Expansion, and (3) Global Relevance Propagation and Evidence Ranking.

\textbf{Phase 1: Initial Anchor Identification}
We extract the explicit entity set $\mathcal{E}_q$ from the query $q$ and map them to the cue nodes $\mathcal{V}_C$ to generate the initial active set $\mathcal{S}_0$, which is defined as:
\begin{equation*}
\mathcal{S}_0 = \{ v \in \mathcal{V}_C \mid \exists e \in \mathcal{E}_q, \text{sim}(\Phi(e), \Phi(v)) \geq \tau_{\text{init}} \},
\end{equation*}
where $\Phi(\cdot)$ denotes the embedding function, $\text{sim}(\cdot, \cdot)$ is the cosine similarity, and $\tau_{\text{init}}$ is the matching threshold.

\textbf{Phase 2: Iterative Context Expansion}
In this phase, HERO emulates the human ``following-the-clues'' recall process: for $K$ iterations, the system traverses a cue--context--cue pathway. Firstly, we  expand $\mathcal{S}_{k-1}$ to include synonymous nodes, handle the lexical diversity situation:
\begin{equation*}
\begin{aligned}
\mathcal{S}'_{k-1} = \{ v \in \mathcal{V}_C \mid {}& \exists u \in \mathcal{S}_{k-1}, \\
& \text{sim}(\Phi(v), \Phi(u)) \geq \tau_{\text{alias}} \},
\end{aligned}
\end{equation*}
where $\tau_{\text{alias}}$ is proposed to constrain semantic drift. We then collect context nodes connected to the current cues $\mathcal{S}'_{k-1}$:
\begin{equation*}
\mathcal{N}_k = \bigcup_{v \in \mathcal{S}'_{k-1}} \text{Adj}(v) \cap (\mathcal{V}_u \cup \mathcal{V}_p),
\end{equation*}
Next, we filter them by query relevance:
\begin{equation*}
\mathcal{N}^*_k = \{ n \in \mathcal{N}_k \mid \text{sim}(\Phi(n), \Phi(q)) \geq \tau_{\text{iter}} \},
\end{equation*}
we activate new cue nodes in $\mathcal{V}_C$ connected to the selected contexts $\mathcal{N}^*_k$ to generate cues:
\begin{equation*}
\mathcal{S}_{k} = \{ c \in \mathcal{V}_C \mid \exists n \in \mathcal{N}^*_k, (n, c) \in \mathcal{E} \} \setminus \bigcup_{i=0}^{k-1} \mathcal{S}_i,
\end{equation*}
for each new cue $c \in \mathcal{S}_k$, set its activation weight by transferring from a previous cue $c' \in \mathcal{S}'_{k-1}$ through the linking context $n \in \mathcal{N}^*_k$:
\begin{equation*}
W(c)_k = W(c')_{k-1} \times \text{sim}(\Phi(n), \Phi(q)).
\end{equation*}

\textbf{Phase 3: Global Relevance Propagation and Evidence Ranking}
After K iterations, the system aggregates the activation signals to identify the most salient episodic traces ($v_t$). The activation intensity of an episodic trace is calculated by combining dense retrieval relevance and the cue-matching bonus:
\begin{equation*}
\begin{aligned}
W(v_t) ={}& \alpha\,\text{sim}(\Phi(v_t), \Phi(q)) \\
&{}+\log\left(1+\sum_{c \in \mathcal{S}_{\text{all}}}
\frac{W(c)\ln(1+\text{f}(c,v_t))}{L_c}\right),
\end{aligned}
\end{equation*}
where $\mathcal{S}_{\text{all}}=\bigcup_{k=0}^{K}\mathcal{S}_k$
denotes the global activated cue set. $W(c)$ is the activation weight of cue node $c$, $\text{f}(c, v_t)$ is its occurrence frequency in trace $v_t$, and $L_c$ is the hierarchical depth of cue node $c$.

Then, node-level scores are integrated into a restart probability vector $\mathbf{r}$ constructed from $\{W(v) \mid v \in \mathcal{V}_C \cup \mathcal{V}_t\}$; $v_p$ is masked during propagation and thus excluded from $\mathbf{r}$. Following~\cite{DBLP:conf/nips/GutierrezS0Y024,DBLP:journals/corr/abs-2510-10114}, we employ Personalized PageRank (PPR)~\cite{haveliwala2002topic} on $G$ with $\mathbf{r}$ as the restart distribution and the Profile Insights nodes $v_p$ masked, which can be viewed as energy propagation on the memory graph:
\begin{equation*}
\mathbf{a} = (1 - d)\,\mathbf{r} + d\,\mathbf{a}\,\mathbf{P}_{\text{mask}},
\end{equation*}
where $\mathbf{a}$ denotes the stationary PPR score vector, $d \in (0, 1)$ is the damping factor, and $\mathbf{P}_{\text{mask}}$ is the row-normalized transition matrix after masking profile nodes $v_p$ (and their incident edges), ensuring that the propagation only flows through evidence-bearing nodes.
Finally, we rank Episodic Traces $v_t$ by their PPR scores, and apply a re-ranking model to return the top-$k$ Episodic Traces  as reasoning evidence for the LLM.

\section{Experiments}
In this section, we evaluate our proposed HERO framework with respect to the following research questions:

\begin{itemize}
\item \textbf{RQ1.} Compared with methods based on compression or rewriting, can HERO provide a more accurate and faithful retrieval of fine-grained facts from long-term memory?

\item \textbf{RQ2.} Can HERO better support reasoning across multiple sessions and events?

\item \textbf{RQ3.} Can HERO capture users’ historical events, traits, and preferences to enable personalized responses?

\item \textbf{RQ4.} How efficient is HERO during retrieval?
\end{itemize}

\begin{table*}
  \centering
  \resizebox{\textwidth}{!}{
  \begin{tabular}{lcccccccccccccccc}
    \toprule
    Method
    & \multicolumn{3}{c}{Multi-hop}
    & \multicolumn{3}{c}{Temporal}
    & \multicolumn{3}{c}{Open-domain}
    & \multicolumn{3}{c}{Single-hop}
    & \multicolumn{3}{c}{Overall Average} \\
    
    \cmidrule(lr){2-4}
    \cmidrule(lr){5-7}
    \cmidrule(lr){8-10}
    \cmidrule(lr){11-13}
    \cmidrule(lr){14-16}
    
    & ACC & F1& BLEU-1
    & ACC & F1 & BLEU-1
    & ACC & F1 & BLEU-1
    & ACC & F1 & BLEU-1
    & ACC & F1 & BLEU-1 \\
    
    \midrule
    
    HERO
    & \textbf{85.11} & \textbf{43.52} & \textbf{32.57}
    & \textbf{86.60} & \textbf{63.63} & \textbf{57.82}
    & \textbf{72.92} & \textbf{34.67} & \textbf{30.13}
    & \textbf{91.20} & \textbf{59.81} & \textbf{52.15}
    & \textbf{87.99} & \textbf{56.06} & \textbf{48.38} \\
    
    CAM
    & 76.60 & 29.89 & 14.72
    & 76.32 & 40.34 & 35.51
    & 66.67 & 19.37 & 15.57
    & 87.40 & 30.70 & 24.55
    & 81.82 & 31.85 & 24.48 \\
    
    
    EverMemOS
    & 81.91 & 41.55 & 28.31
    & 83.80 & 59.08 & 52.09
    & 69.79 & 25.76 & 19.33
    & 86.44 & 43.78 & 36.06
    & 84.03 & 45.48 & 36.94 \\
    
    LinearRAG
    & 52.13 & 28.83 & 21.11
    & 53.27 & 41.50 & 36.82
    & 55.21 & 22.44 & 16.73
    & 66.35 & 44.70 & 39.24
    & 60.32 & 39.74 & 34.01 \\


    MemoryOS
    & 57.45 & 35.11 & 26.62
    & 44.86 & 32.05 & 26.68
    & 53.13 & 23.57 & 12.77
    & 74.67 & 52.10 & 45.47
    & 63.96 & 43.03 & 36.06 \\

    Mem0
    & 67.02 & 34.99 & 24.07
    & 40.19 & 32.88 & 29.02
    & 63.54 & 21.83 & 15.98
    & 73.48 & 39.79 & 33.30
    & 64.74 & 36.35 & 29.64 \\
    \bottomrule
  \end{tabular}}
  \caption{Performance Comparison on the LoCoMo benchmark. All values represent the performance metric for each task (e.g., accuracy \%). Best results are highlighted in bold.}
    \label{tab:LoCoMo_results}
\end{table*}

\begin{table}[t]
\centering
\begin{tabular}{lcc}
\toprule
\textbf{Category} & \textbf{R@5} & \textbf{R@10} \\
\midrule
Single-hop   & 0.935 & 0.960 \\
Multi-hop    & 0.915 & 0.947 \\
Temporal     & 0.935 & 0.953 \\
Open-domain  & 0.604 & 0.667 \\
\midrule
Overall      & \textbf{0.910} & \textbf{0.938} \\
\bottomrule
\end{tabular}
\caption{Retrieval quality on the LoCoMo benchmark. }
\label{tab:retrieval_quality}
\end{table}

\begin{table*}[t]
\centering
  \small
  \begin{tabular}{%
    p{0.22\textwidth}
    p{0.42\textwidth}
    p{0.31\textwidth}}
    \toprule
    \textbf{Question \& Gold Answer} &
    \textbf{Original Context} &
    \textbf{Generated Memory} \\
    \midrule
    \textbf{Q:} What did Evan start painting due to a friend's gift?
    \textbf{A:} \textbf{forest scene}
    &
    \textit{    Evan shared a photo of a painting of a forest scene on an easel, and said,
    ``Thanks, Sam! It all started when a friend of mine gave me this painting one day,
    it inspired me a lot and that's when I started painting...''}
    &
    \textit{Memory Content: Evan thanked Sam and mentioned starting painting classes a few days ago.
    Evan shared a forest painting, inspired by a gift from a friend.} \\
    \midrule
    \textbf{Q:} How often does Audrey take her pups to the park for practice?
    \textbf{A:} \textbf{Twice a week}
    &
    \textit{Audrey shared a photo of a group of dogs sitting on a dirt road and said,    ``...it's a great physical and mental workout. I take them to the park twice a week
    for practice---it's been a great bonding experience...''}
    &
    \textit{Memory Content: Audrey highlighted her twice-weekly agility classes and the positive impact
    on her dogs' social and physical development.} \\
    \bottomrule
  \end{tabular}
  \caption{Case study comparing original context and generated memory on the LoCoMo benchmark.}
\label{tab:case_study_memory}
\end{table*}

\subsection{Experimental Setup}


\textbf{Datasets and Metrics.}
We evaluate HERO on two benchmarks. LoCoMo~\cite{LoCoMo} contains 10 long-term
conversations with about 600 turns, 16K tokens, and up to 32 sessions each,
covering single-hop, multi-hop, temporal, and open-domain QA; we report F1, BLEU-1, and LLM-judged accuracy following the standard
protocol~\cite{journals/corr/abs-2504-19413,2026arXiv260102163H}.
PERSONAMEM-32k~\cite{PERSONAMEM} evaluates personalized memory over 180 simulated
user--agent histories across real-world scenarios, covering user facts, evolving
preferences, update reasons, and preference-aligned recommendations; we report
multiple-choice accuracy.

\textbf{Baselines.}
We compare HERO with state-of-the-art baselines. Mem0\cite{journals/corr/abs-2504-19413}, MemoryOS\cite{Memoryos} and EverMemOS\cite{2026arXiv260102163H} rely on LLM-based rewriting or memory updates to transform raw dialogue into compact memory representations. CAM\cite{DBLP:journals/corr/abs-2510-05520} builds hierarchical summary nodes through topic clustering. LinearRAG\cite{DBLP:journals/corr/abs-2510-10114} is a lightweight graph retrieval method.

\textbf{Implementation Details.}
For the main experiments, methods are evaluated using Qwen2.5-72B-Instruct~\cite{qwen2025qwen25technicalreport} as the backbone LLM and Qwen3-Embedding-8B~\cite{qwen3embedding} as the embedding model. Mem0 is evaluated using its official hosted version in the main comparison. All methods share the same answer-generation prompts, as detailed in the Appendix.








\subsection{Overall Performance}

\textbf{RQ1. HERO achieves leading response correctness and fidelity by preserving raw episodic context.}
We compare HERO with baselines on the LoCoMo dataset. The four LoCoMo task types focus on fact-based question answering, where answers must be retrieved from the dialogue history or inferred by combining multiple clues. As shown in Table~\ref{tab:LoCoMo_results}, HERO achieves the best overall performance across all metrics, reaching 87.99\% accuracy, 56.06\% F1, and 48.38\% BLEU-1. These results indicate that HERO not only produces correct answers, but also stays closer to the source dialogue text in wording, because its generation is directly grounded in raw text.

To further verify whether the gain comes from better evidence retrieval, we
evaluate retrieval quality on LoCoMo using Recall@5 and Recall@10
against annotated supporting spans. As shown in Table~\ref{tab:retrieval_quality},
HERO achieves an overall Recall@10 of 0.938, indicating that the graph-based
retriever can locate the relevant raw evidence in most cases. The lower recall on open-domain questions is expected, since such questions
often require external commonsense or world knowledge not fully expressed in
the dialogue history.

These results highlight the importance of memory storage strategy. Methods
based on LLM rewriting or summarization perform irreversible compression before
the downstream query is known, which may discard details or introduce semantic
drift. This is consistent with prior analyses of LLM-generated event summaries,
which report missing temporal or causal links, hallucinated details, and
incorrect speaker attributions~\cite{LoCoMo}. The case study in
Table~\ref{tab:case_study_memory} further illustrates this issue: generated
memories may preserve surface keywords while distorting causal links or event
contexts.


\begin{table*}
  \centering
  \resizebox{\textwidth}{!}{
  \begin{tabular}{lcccccccc}
    \toprule
    Method
    & \shortstack{Generalize \\ to New\\Scenarios}
    & \shortstack{Provide \\Preference-Aligned\\Recommendations}
    & \shortstack{Recall \\User\\Shared Facts}
    & \shortstack{Acknowledge \\Latest\\User Preferences}
    & \shortstack{Revisit \\Reasons behind\\Preference Updates}
    & \shortstack{Suggest \\New\\Ideas}
    & \shortstack{Track \\ Full Preference\\Evolution}
    & \shortstack{Overall\\Avg} \\
    
    \midrule
    
    HERO
    & \textbf{82.46}
    & \textbf{80.00}
    & 68.99
    &  \textbf{76.47}
    & \textbf{87.88}
    & \textbf{37.63}
    & \textbf{72.66}
    & \textbf{70.63} \\

    CAM
    & 70.18
    & 54.55
    & 65.89
    & \textbf{76.47}
    & 75.76
    & 36.56
    & 61.87
    & 61.63 \\

    Mem0
    & 73.68
    & 72.73
    & 61.24
    & 70.59
    & 79.80
    & 35.48
    & 68.34
    & 64.52 \\

    EverMemOS
    & 80.70
    & 72.73
    & 68.99
    & \textbf{76.47}
    & 83.84
    & 22.58
    & 71.22
    & 66.38 \\
    

    LinearRAG
    & 68.42
    & 74.55
    & \textbf{72.09}
    & 70.59
    & 81.82
    & 21.51
    & 71.94
    & 65.35 \\

    MemoryOS
    & \textbf{82.46}
    & 67.27
    & 65.12
    & 70.59
    & 86.87
    & 22.58
    & 71.94
    & 65.70 \\
    
    \bottomrule
  \end{tabular}}
    \caption{Performance comparison on the PERSONAMEM benchmark. All values represent the performance metric for each task. Best results are highlighted in bold.}
  \label{tab:PERSONAMEM_results}
\end{table*}

\textbf{RQ2. HERO demonstrates cross-session reasoning capabilities.}
As shown in Table~\ref{tab:LoCoMo_results}, HERO achieves an accuracy of 85.11\% on the Multi-hop tasks, which require linking scattered clues across multiple
sessions. This result indicates that HERO can recover long-range dependencies
rather than relying only on locally similar dialogue snippets.
For example, in a LoCoMo instance where the query asks
\texttt{``Who did John go to yoga with?''}, the relevant evidence is split
across two temporally distant sessions. In an earlier dialogue on
\texttt{2023-02-25}, John stated that \texttt{``my colleague Rob invited me to
a beginner's yoga class''}; in a later session on \texttt{2023-04-07}, he
mentioned that he \texttt{``started a weekend yoga class with a colleague''}
without naming the person. Neither utterance alone is sufficient to answer the
query. HERO bridges them by following cue-based links from
\texttt{yoga class} to \texttt{colleague} and then to \texttt{Rob}, thereby
retrieving the cross-session evidence needed to infer the answer. This example illustrates how HERO's episodic graph supports multi-hop
retrieval: HERO connects episodic traces through entity cues, allowing retrieval to follow the graph topology and perform multi-hop walks, which better captures long-range dependencies across sessions.

    
    
    
    
    
    
    

\begin{figure*}[t]
  \centering
  \includegraphics[width=\linewidth]{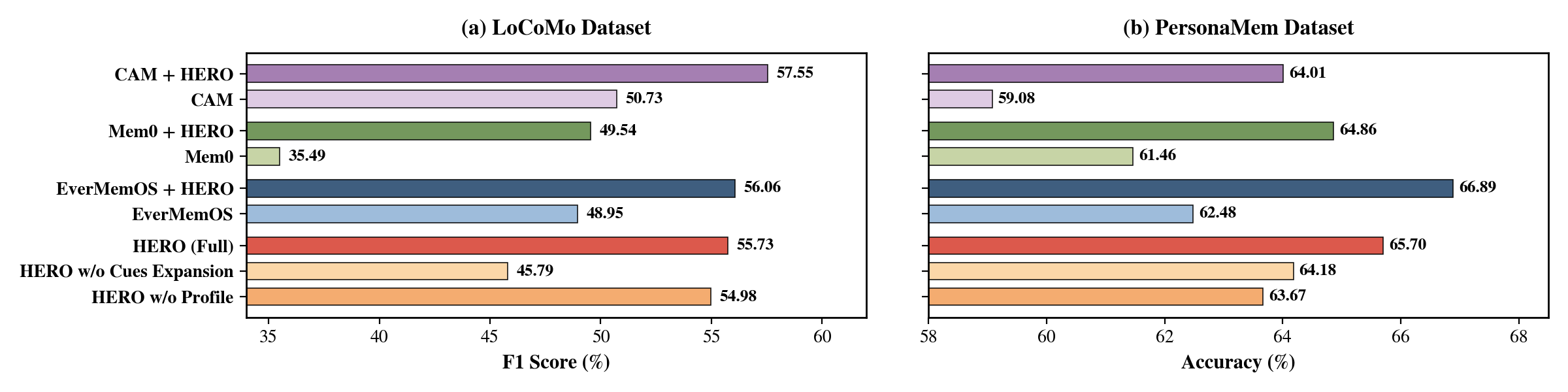}
  \caption{Ablation Studies of HERO. }
  \label{fig:ablation}
\end{figure*}

\textbf{RQ3. HERO is able to capture dynamic human profile and evolving preferences.}
We evaluate HERO on PERSONAMEM, which focuses on tracking user facts, evolving
preferences, and preference-aligned responses across long-term interactions. As shown in Table~\ref{tab:PERSONAMEM_results}, HERO achieves the best overall
accuracy of 70.63\%, outperforming all baselines.

HERO shows clear advantages on preference-oriented categories, including
Preference-Aligned Recommendations (80.00\%), Revisit Reasons behind Preference
Updates (87.88\%), and Track Full Preference Evolution (72.66\%). These results
suggest that the profile layer helps identify user-specific cues that are not
always explicit in the current query. Importantly, profiles in HERO are used as
retrieval guidance rather than final evidence: they steer the retriever toward
relevant raw dialogue traces, while the answer is generated from the retrieved
original snippets. The ablation results in Figure~\ref{fig:ablation} further support this design:
removing the profile layer decreases PERSONAMEM accuracy from 65.70\% to
63.67\%, indicating that profile-enhanced retrieval provides complementary
signals beyond query-only matching.
This is consistent with the finding in~\cite{PERSONAMEM} that relevant raw
messages are important for personalized reasoning.

\begin{figure}[t]
  \centering
  \includegraphics[width=0.85\linewidth]{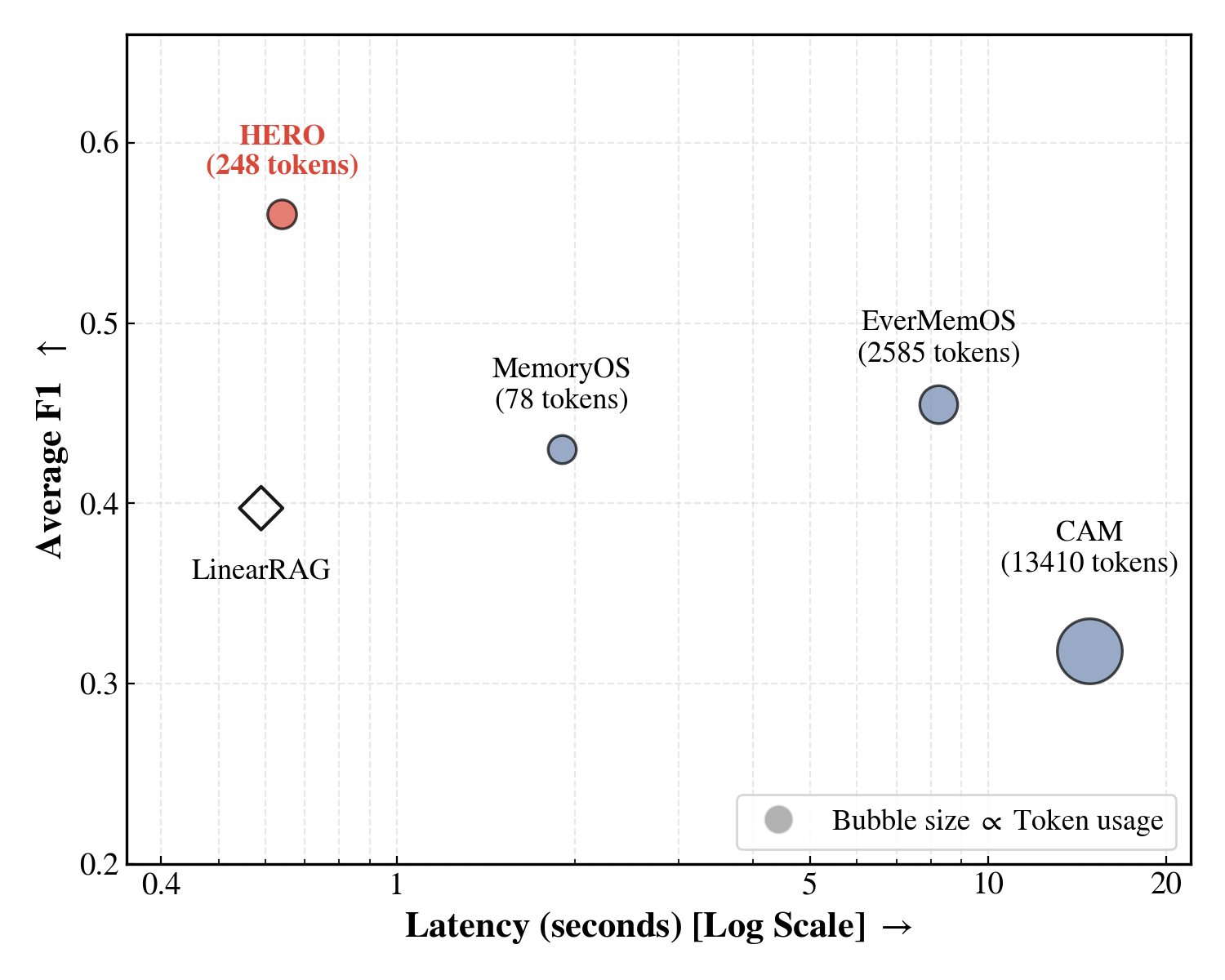}
  \caption{Performance--efficiency trade-off during retrieval on the LoCoMo benchmark. }
  \label{fig:effiency}
\end{figure}

\subsection{Trade-off Analysis}

\textbf{RQ4. HERO achieves an optimal trade-off between system efficiency and response performance.} 
We further analyze the LLM token cost and time efficiency during retrieval on the LoCoMo benchmark. All efficiency metrics (latency and token usage) are reported as the average value per query. Figure~\ref{fig:effiency} plots different methods along three dimensions: retrieval latency (x-axis, in log scale), F1 score (y-axis), and token consumption (bubble size). The bubble size represents the per-query LLM token consumption, including both input and output tokens. 

HERO achieves high effectiveness with low retrieval overhead. It obtains the
highest F1 while using only 248 tokens on average, and its latency remains below
one second. This efficiency comes from HERO's design: heavy semantic linking
and structure building are moved to the memory construction and profile
maintenance stages. During retrieval, HERO only performs lightweight cue
extraction followed by graph traversal, avoiding long chains of online LLM
reasoning.

\subsection{Ablation Study}

To validate the effectiveness of HERO's components, we conduct an ablation study with two variants: (1) \textbf{HERO w/o Profile}, where the profile layer is deactivated to assess the impact of profile-enhanced retrieval; and (2) \textbf{HERO w/o Cues Expansion}, where iterative context expansion is disabled to evaluate multi-hop evidence retrieval. The results in Figure~\ref{fig:ablation}  demonstrate the importance of both components, with different effects across tasks. 
We also use HERO-retrieved raw dialogue segments as auxiliary evidence for
existing baselines, which consistently improves their performance, further
indicating the value of faithful raw-memory retrieval.

\subsection{Hyperparameter Analysis}
We examine whether HERO requires careful hyperparameter tuning. As shown in
Table~\ref{tab:sensitivity}, LoCoMo accuracy remains within 0.801--0.829 and
F1 within 0.546--0.560 when varying the alias matching threshold, retrieval
filtering threshold, and retrieval top-$k$. These results show that HERO is
stable under moderate hyperparameter changes and does not rely on delicate
threshold tuning.

\begin{table}[t]
\centering
\small
\begin{tabular}{llcc}
\toprule
\textbf{Hyperparameter} & \textbf{Value} & \textbf{Acc} & \textbf{F1} \\
\midrule
$\tau_{\text{alias}}$ 
& 0.6  & 0.818 & 0.552 \\
& 0.7  & 0.819 & 0.548 \\
& 0.8* & 0.816 & 0.549 \\
& 0.9  & 0.816 & 0.546 \\
\midrule
$\tau_{\text{init}}=\tau_{\text{iter}}$
& 0.3  & 0.801 & 0.548 \\
& 0.4* & 0.829 & 0.558 \\
& 0.5  & 0.814 & 0.557 \\
& 0.6  & 0.823 & 0.548 \\
\midrule
Top-$k$
& 5   & 0.814 & 0.556 \\
& 10* & 0.828 & 0.560 \\
& 20  & 0.823 & 0.550 \\
& 40  & 0.818 & 0.555 \\
\bottomrule
\end{tabular}
\caption{Hyperparameter sensitivity on LoCoMo. Default values are marked with *.}
\label{tab:sensitivity}
\end{table}


\section{Conclusion}
Long-term memory is a fundamental challenge for building personalized and coherent conversational agents. To empower episodic long-term memory for this task, we propose HERO, a novel human-profile enhanced retrieval optimization framework. HERO enables factually faithful reasoning through an episodic-centered heterogeneous memory graph and an iterative dual-cue activation mechanism, significantly improving personalized responses. Experiments on two benchmark datasets demonstrate encouraging performance gains, accurate retrieval capabilities, and robust cross-session reasoning. These results validate the effectiveness of our framework and suggest that preserving raw episodic context with structured retrieval is a promising direction for long-term agent memory.

\section*{Limitations}

While HERO achieves promising results on long-term factual and personalized
memory tasks, we acknowledge several areas for future exploration. First,
although HERO organizes raw episodic traces through temporal chaining and
semantic cue links, the current implementation does not include an explicit
memory lifecycle mechanism, such as time-aware decay, stale-trace pruning, or
user-controlled deletion. Future work could explore incorporating these
strategies as lightweight local graph updates for longer-running deployments.
Second, HERO relies on LLMs to extract cues and profile insights. These profiles guide retrieval, but are not directly used as evidence for generation, mitigating the risk of grounding responses in noisy abstractions; however, inaccurate  profile cues may still affect the retrieval trajectory. Future work could introduce confidence calibration, and timestamp-aware profile updates to further improve robustness.

\section*{Ethical Considerations}

This work studies long-term memory for conversational agents, which involves 
storing and retrieving user dialogue history and personal profiles. 
All datasets used in this work are publicly released for research purposes 
and do not require additional ethical approval. For real-world deployment, 
we recommend that systems built on HERO adopt explicit data governance 
policies, including user-controlled memory deletion and access transparency, 
consistent with applicable data protection regulations.

We also note that inaccuracies in stored or retrieved memory could lead to 
misleading responses. HERO mitigates this risk by grounding generation in 
raw, unmodified dialogue text rather than LLM-rewritten summaries, reducing 
the chance of introducing hallucinated or distorted information.

\bibliography{main}

@inproceedings{ZhongGGYW24,
  author       = {Wanjun Zhong and
                  Lianghong Guo and
                  Qiqi Gao and
                  He Ye and
                  Yanlin Wang},
  editor       = {Michael J. Wooldridge and
                  Jennifer G. Dy and
                  Sriraam Natarajan},
  title        = {MemoryBank: Enhancing Large Language Models with Long-Term Memory},
  booktitle    = {Thirty-Eighth {AAAI} Conference on Artificial Intelligence, {AAAI}
                  2024, Thirty-Sixth Conference on Innovative Applications of Artificial
                  Intelligence, {IAAI} 2024, Fourteenth Symposium on Educational Advances
                  in Artificial Intelligence, {EAAI} 2014, February 20-27, 2024, Vancouver,
                  Canada},
  pages        = {19724--19731},
  publisher    = {{AAAI} Press},
  year         = {2024},
  url          = {https://doi.org/10.1609/aaai.v38i17.29946},
  doi          = {10.1609/AAAI.V38I17.29946},
  bibsource    = {dblp computer science bibliography, https://dblp.org}
}

@article{DBLP:journals/corr/abs-2505-22101,
  author       = {Zhiyu Li and
                  Shichao Song and
                  Hanyu Wang and
                  Simin Niu and
                  Ding Chen and
                  Jiawei Yang and
                  Chenyang Xi and
                  Huayi Lai and
                  Jihao Zhao and
                  Yezhaohui Wang and
                  Junpeng Ren and
                  Zehao Lin and
                  Jiahao Huo and
                  Tianyi Chen and
                  Kai Chen and
                  Kehang Li and
                  Zhiqiang Yin and
                  Qingchen Yu and
                  Bo Tang and
                  Hongkang Yang and
                  Zhi{-}Qin John Xu and
                  Feiyu Xiong},
  title        = {MemOS: An Operating System for Memory-Augmented Generation {(MAG)}
                  in Large Language Models},
  journal      = {CoRR},
  volume       = {abs/2505.22101},
  year         = {2025},
  url          = {https://doi.org/10.48550/arXiv.2505.22101},
  doi          = {10.48550/ARXIV.2505.22101},
  eprinttype    = {arXiv},
  eprint       = {2505.22101},
  bibsource    = {dblp computer science bibliography, https://dblp.org}
}

@article{journals/corr/abs-2510-18866,
  author       = {Jizhan Fang and
                  Xinle Deng and
                  Haoming Xu and
                  Ziyan Jiang and
                  Yuqi Tang and
                  Ziwen Xu and
                  Shumin Deng and
                  Yunzhi Yao and
                  Mengru Wang and
                  Shuofei Qiao and
                  Huajun Chen and
                  Ningyu Zhang},
  title        = {LightMem: Lightweight and Efficient Memory-Augmented Generation},
  journal      = {CoRR},
  volume       = {abs/2510.18866},
  year         = {2025},
  url          = {https://doi.org/10.48550/arXiv.2510.18866},
  doi          = {10.48550/ARXIV.2510.18866},
  eprinttype    = {arXiv},
  eprint       = {2510.18866},
  bibsource    = {dblp computer science bibliography, https://dblp.org}}

@inproceedings{DBLP:conf/iclr/PanWJLCL0LZQ025,
  author       = {Zhuoshi Pan and
                  Qianhui Wu and
                  Huiqiang Jiang and
                  Xufang Luo and
                  Hao Cheng and
                  Dongsheng Li and
                  Yuqing Yang and
                  Chin{-}Yew Lin and
                  H. Vicky Zhao and
                  Lili Qiu and
                  Jianfeng Gao},
  title        = {SeCom: On Memory Construction and Retrieval for Personalized Conversational
                  Agents},
  booktitle    = {The Thirteenth International Conference on Learning Representations,
                  {ICLR} 2025, Singapore, April 24-28, 2025},
  publisher    = {OpenReview.net},
  year         = {2025},
  url          = {https://openreview.net/forum?id=xKDZAW0He3},
  bibsource    = {dblp computer science bibliography, https://dblp.org}
}

@inproceedings{DBLP:conf/naacl/LiYZDWC25,
  author       = {Hao Li and
                  Chenghao Yang and
                  An Zhang and
                  Yang Deng and
                  Xiang Wang and
                  Tat{-}Seng Chua},
  editor       = {Luis Chiruzzo and
                  Alan Ritter and
                  Lu Wang},
  title        = {Hello Again! LLM-powered Personalized Agent for Long-term Dialogue},
  booktitle    = {Proceedings of the 2025 Conference of the Nations of the Americas
                  Chapter of the Association for Computational Linguistics: Human Language
                  Technologies, {NAACL} 2025 - Volume 1: Long Papers, Albuquerque, New
                  Mexico, USA, April 29 - May 4, 2025},
  pages        = {5259--5276},
  publisher    = {Association for Computational Linguistics},
  year         = {2025},
  url          = {https://doi.org/10.18653/v1/2025.naacl-long.272},
  doi          = {10.18653/V1/2025.NAACL-LONG.272},
  bibsource    = {dblp computer science bibliography, https://dblp.org}
}

@inproceedings{DBLP:conf/icml/LeeCFCF24,
  author       = {Kuang{-}Huei Lee and
                  Xinyun Chen and
                  Hiroki Furuta and
                  John F. Canny and
                  Ian Fischer},
  title        = {A Human-Inspired Reading Agent with Gist Memory of Very Long Contexts},
  booktitle    = {Forty-first International Conference on Machine Learning, {ICML} 2024,
                  Vienna, Austria, July 21-27, 2024},
  publisher    = {OpenReview.net},
  year         = {2024},
  url          = {https://openreview.net/forum?id=OTmcsyEO5G},
  bibsource    = {dblp computer science bibliography, https://dblp.org}
}

@inproceedings{DBLP:conf/emnlp/00110PCML0024,
  author       = {Wenhao Yu and
                  Hongming Zhang and
                  Xiaoman Pan and
                  Peixin Cao and
                  Kaixin Ma and
                  Jian Li and
                  Hongwei Wang and
                  Dong Yu},
  editor       = {Yaser Al{-}Onaizan and
                  Mohit Bansal and
                  Yun{-}Nung Chen},
  title        = {Chain-of-Note: Enhancing Robustness in Retrieval-Augmented Language
                  Models},
  booktitle    = {Proceedings of the 2024 Conference on Empirical Methods in Natural
                  Language Processing, {EMNLP} 2024, Miami, FL, USA, November 12-16,
                  2024},
  pages        = {14672--14685},
  publisher    = {Association for Computational Linguistics},
  year         = {2024},
  url          = {https://doi.org/10.18653/v1/2024.emnlp-main.813},
  doi          = {10.18653/V1/2024.EMNLP-MAIN.813},
  bibsource    = {dblp computer science bibliography, https://dblp.org}}

@article{DBLP:journals/corr/abs-2506-06326,
  author       = {Jiazheng Kang and
                  Mingming Ji and
                  Zhe Zhao and
                  Ting Bai},
  title        = {Memory {OS} of {AI} Agent},
  journal      = {CoRR},
  volume       = {abs/2506.06326},
  year         = {2025},
  url          = {https://doi.org/10.48550/arXiv.2506.06326},
  doi          = {10.48550/ARXIV.2506.06326},
  eprinttype    = {arXiv},
  eprint       = {2506.06326},
  bibsource    = {dblp computer science bibliography, https://dblp.org}
}

@article{journals/corr/abs-2504-19413,
  author       = {Prateek Chhikara and
                  Dev Khant and
                  Saket Aryan and
                  Taranjeet Singh and
                  Deshraj Yadav},
  title        = {Mem0: Building Production-Ready {AI} Agents with Scalable Long-Term
                  Memory},
  journal      = {CoRR},
  volume       = {abs/2504.19413},
  year         = {2025},
  url          = {https://doi.org/10.48550/arXiv.2504.19413},
  doi          = {10.48550/ARXIV.2504.19413},
  eprinttype    = {arXiv},
  eprint       = {2504.19413},
  bibsource    = {dblp computer science bibliography, https://dblp.org}
}

@article{LangChain,
  author       = {LangChain Team},
  title        = {Conversational rag},
  url          = {. https://python.langchain.com/v0.2/docs/t utorials/qa_chat_history/},
  year = 2023
}

@ARTICLE{2026arXiv260102163H,
       author = {{Hu}, Chuanrui and {Gao}, Xingze and {Zhou}, Zuyi and {Xu}, Dannong and {Bai}, Yi and {Li}, Xintong and {Zhang}, Hui and {Li}, Tong and {Zhang}, Chong and {Bing}, Lidong and {Deng}, Yafeng},
        title = "{EverMemOS: A Self-Organizing Memory Operating System for Structured Long-Horizon Reasoning}",
      journal = {arXiv e-prints},
         year = 2026,
        month = jan,
          eid = {arXiv:2601.02163},
        pages = {arXiv:2601.02163},
          doi = {10.48550/arXiv.2601.02163},
archivePrefix = {arXiv},
       eprint = {2601.02163},
 primaryClass = {cs.AI},
       adsurl = {https://ui.adsabs.harvard.edu/abs/2026arXiv260102163H}
}

@article{DBLP:journals/corr/abs-2308-08239,
  author       = {Junru Lu and
                  Siyu An and
                  Mingbao Lin and
                  Gabriele Pergola and
                  Yulan He and
                  Di Yin and
                  Xing Sun and
                  Yunsheng Wu},
  title        = {MemoChat: Tuning LLMs to Use Memos for Consistent Long-Range Open-Domain
                  Conversation},
  journal      = {CoRR},
  volume       = {abs/2308.08239},
  year         = {2023},
  url          = {https://doi.org/10.48550/arXiv.2308.08239},
  doi          = {10.48550/ARXIV.2308.08239},
  eprinttype    = {arXiv},
  eprint       = {2308.08239},
  bibsource    = {dblp computer science bibliography, https://dblp.org}
}

@inproceedings{DBLP:conf/nips/LewisPPPKGKLYR020,
  author       = {Patrick Lewis and
                  Ethan Perez and
                  Aleksandra Piktus and
                  Fabio Petroni and
                  Vladimir Karpukhin and
                  Naman Goyal and
                  Heinrich K{\"{u}}ttler and
                  Mike Lewis and
                  Wen{-}tau Yih and
                  Tim Rockt{\"{a}}schel and
                  Sebastian Riedel and
                  Douwe Kiela},
  editor       = {Hugo Larochelle and
                  Marc'Aurelio Ranzato and
                  Raia Hadsell and
                  Maria{-}Florina Balcan and
                  Hsuan{-}Tien Lin},
  title        = {Retrieval-Augmented Generation for Knowledge-Intensive {NLP} Tasks},
  booktitle    = {Advances in Neural Information Processing Systems 33: Annual Conference
                  on Neural Information Processing Systems 2020, NeurIPS 2020, December
                  6-12, 2020, virtual},
  year         = {2020},
  url          = {https://proceedings.neurips.cc/paper/2020/hash/6b493230205f780e1bc26945df7481e5-Abstract.html},
  bibsource    = {dblp computer science bibliography, https://dblp.org}
}

@article{DBLP:journals/corr/abs-2310-08560,
  author       = {Charles Packer and
                  Vivian Fang and
                  Shishir G. Patil and
                  Kevin Lin and
                  Sarah Wooders and
                  Joseph E. Gonzalez},
  title        = {MemGPT: Towards LLMs as Operating Systems},
  journal      = {CoRR},
  volume       = {abs/2310.08560},
  year         = {2023},
  url          = {https://doi.org/10.48550/arXiv.2310.08560},
  doi          = {10.48550/ARXIV.2310.08560},
  eprinttype    = {arXiv},
  eprint       = {2310.08560},
  bibsource    = {dblp computer science bibliography, https://dblp.org}
}

@article{DBLP:journals/corr/abs-2501-13956,
  author       = {Preston Rasmussen and
                  Pavlo Paliychuk and
                  Travis Beauvais and
                  Jack Ryan and
                  Daniel Chalef},
  title        = {Zep: {A} Temporal Knowledge Graph Architecture for Agent Memory},
  journal      = {CoRR},
  volume       = {abs/2501.13956},
  year         = {2025},
  url          = {https://doi.org/10.48550/arXiv.2501.13956},
  doi          = {10.48550/ARXIV.2501.13956},
  eprinttype    = {arXiv},
  eprint       = {2501.13956},
  bibsource    = {dblp computer science bibliography, https://dblp.org}
}

@article{DBLP:journals/corr/abs-2510-10114,
  author       = {Luyao Zhuang and
                  Shengyuan Chen and
                  Yilin Xiao and
                  Huachi Zhou and
                  Yujing Zhang and
                  Hao Chen and
                  Qinggang Zhang and
                  Xiao Huang},
  title        = {LinearRAG: Linear Graph Retrieval Augmented Generation on Large-scale
                  Corpora},
  journal      = {CoRR},
  volume       = {abs/2510.10114},
  year         = {2025},
  url          = {https://doi.org/10.48550/arXiv.2510.10114},
  doi          = {10.48550/ARXIV.2510.10114},
  eprinttype    = {arXiv},
  eprint       = {2510.10114},
  bibsource    = {dblp computer science bibliography, https://dblp.org}
}

@article{DBLP:journals/corr/abs-2502-12110,
  title={A-mem: Agentic memory for llm agents},
  author={Xu, Wujiang and Liang, Zujie and Mei, Kai and Gao, Hang and Tan, Juntao and Zhang, Yongfeng},
  journal={Advances in Neural Information Processing Systems},
  volume={38},
  pages={17577--17604},
  year={2026}
}

@article{DBLP:journals/corr/abs-2410-05779,
  author       = {Zirui Guo and
                  Lianghao Xia and
                  Yanhua Yu and
                  Tu Ao and
                  Chao Huang},
  title        = {LightRAG: Simple and Fast Retrieval-Augmented Generation},
  journal      = {CoRR},
  volume       = {abs/2410.05779},
  year         = {2024},
  url          = {https://doi.org/10.48550/arXiv.2410.05779},
  doi          = {10.48550/ARXIV.2410.05779},
  eprinttype    = {arXiv},
  eprint       = {2410.05779},
  bibsource    = {dblp computer science bibliography, https://dblp.org}
}

@article{DBLP:journals/corr/abs-2404-16130,
  author       = {Darren Edge and
                  Ha Trinh and
                  Newman Cheng and
                  Joshua Bradley and
                  Alex Chao and
                  Apurva Mody and
                  Steven Truitt and
                  Jonathan Larson},
  title        = {From Local to Global: {A} Graph {RAG} Approach to Query-Focused Summarization},
  journal      = {CoRR},
  volume       = {abs/2404.16130},
  year         = {2024},
  url          = {https://doi.org/10.48550/arXiv.2404.16130},
  doi          = {10.48550/ARXIV.2404.16130},
  eprinttype    = {arXiv},
  eprint       = {2404.16130},
  bibsource    = {dblp computer science bibliography, https://dblp.org}
}

@article{DBLP:journals/corr/abs-2501-00309,
  author       = {Haoyu Han and
                  Yu Wang and
                  Harry Shomer and
                  Kai Guo and
                  Jiayuan Ding and
                  Yongjia Lei and
                  Mahantesh Halappanavar and
                  Ryan A. Rossi and
                  Subhabrata Mukherjee and
                  Xianfeng Tang and
                  Qi He and
                  Zhigang Hua and
                  Bo Long and
                  Tong Zhao and
                  Neil Shah and
                  Amin Javari and
                  Yinglong Xia and
                  Jiliang Tang},
  title        = {Retrieval-Augmented Generation with Graphs (GraphRAG)},
  journal      = {CoRR},
  volume       = {abs/2501.00309},
  year         = {2025},
  url          = {https://doi.org/10.48550/arXiv.2501.00309},
  doi          = {10.48550/ARXIV.2501.00309},
  eprinttype    = {arXiv},
  eprint       = {2501.00309},
  bibsource    = {dblp computer science bibliography, https://dblp.org}
}

@inproceedings{DBLP:conf/iclr/SarthiATKGM24,
  author       = {Parth Sarthi and
                  Salman Abdullah and
                  Aditi Tuli and
                  Shubh Khanna and
                  Anna Goldie and
                  Christopher D. Manning},
  title        = {{RAPTOR:} Recursive Abstractive Processing for Tree-Organized Retrieval},
  booktitle    = {The Twelfth International Conference on Learning Representations,
                  {ICLR} 2024, Vienna, Austria, May 7-11, 2024},
  publisher    = {OpenReview.net},
  year         = {2024},
  url          = {https://openreview.net/forum?id=GN921JHCRw},
  bibsource    = {dblp computer science bibliography, https://dblp.org}
}

@inproceedings{DBLP:conf/iclr/RezazadehLWB25,
  author       = {Alireza Rezazadeh and
                  Zichao Li and
                  Wei Wei and
                  Yujia Bao},
  title        = {From Isolated Conversations to Hierarchical Schemas: Dynamic Tree
                  Memory Representation for LLMs},
  booktitle    = {The Thirteenth International Conference on Learning Representations,
                  {ICLR} 2025, Singapore, April 24-28, 2025},
  publisher    = {OpenReview.net},
  year         = {2025},
  url          = {https://openreview.net/forum?id=moXtEmCleY},
  bibsource    = {dblp computer science bibliography, https://dblp.org}
}

@inproceedings{DBLP:conf/ijcai/AnokhinSSEK0B25,
  author       = {Petr Anokhin and
                  Nikita Semenov and
                  Artyom Y. Sorokin and
                  Dmitry Evseev and
                  Andrey Kravchenko and
                  Mikhail Burtsev and
                  Evgeny Burnaev},
  title        = {AriGraph: Learning Knowledge Graph World Models with Episodic Memory
                  for {LLM} Agents},
  booktitle    = {Proceedings of the Thirty-Fourth International Joint Conference on
                  Artificial Intelligence, {IJCAI} 2025, Montreal, Canada, August 16-22,
                  2025},
  pages        = {12--20},
  publisher    = {ijcai.org},
  year         = {2025},
  url          = {https://doi.org/10.24963/ijcai.2025/2},
  doi          = {10.24963/IJCAI.2025/2},
  bibsource    = {dblp computer science bibliography, https://dblp.org}
}

@article{DBLP:journals/corr/abs-2506-07398,
  author       = {Guibin Zhang and
                  Muxin Fu and
                  Guancheng Wan and
                  Miao Yu and
                  Kun Wang and
                  Shuicheng Yan},
  title        = {G-Memory: Tracing Hierarchical Memory for Multi-Agent Systems},
  journal      = {CoRR},
  volume       = {abs/2506.07398},
  year         = {2025},
  url          = {https://doi.org/10.48550/arXiv.2506.07398},
  doi          = {10.48550/ARXIV.2506.07398},
  eprinttype    = {arXiv},
  eprint       = {2506.07398},
  bibsource    = {dblp computer science bibliography, https://dblp.org}
}

@inproceedings{DBLP:conf/nips/GutierrezS0Y024,
  author       = {Bernal Jimenez Gutierrez and
                  Yiheng Shu and
                  Yu Gu and
                  Michihiro Yasunaga and
                  Yu Su},
  editor       = {Amir Globersons and
                  Lester Mackey and
                  Danielle Belgrave and
                  Angela Fan and
                  Ulrich Paquet and
                  Jakub M. Tomczak and
                  Cheng Zhang},
  title        = {HippoRAG: Neurobiologically Inspired Long-Term Memory for Large Language
                  Models},
  booktitle    = {Advances in Neural Information Processing Systems 38: Annual Conference
                  on Neural Information Processing Systems 2024, NeurIPS 2024, Vancouver,
                  BC, Canada, December 10 - 15, 2024},
  year         = {2024},
  url          = {http://papers.nips.cc/paper\_files/paper/2024/hash/6ddc001d07ca4f319af96a3024f6dbd1-Abstract-Conference.html},
  bibsource    = {dblp computer science bibliography, https://dblp.org}
}

@article{DBLP:journals/corr/abs-2510-05520,
  title={Cam: A constructivist view of agentic memory for llm-based reading comprehension},
  author={Li, Rui and Zhang, Zeyu and Bo, Xiaohe and Tian, Zihang and Chen, Xu and Dai, Quanyu and Dong, Zhenhua and Tang, Ruiming},
  journal={Advances in Neural Information Processing Systems},
  volume={38},
  pages={113381--113406},
  year={2026}
}

@inproceedings{locomo,
  title={Evaluating very long-term conversational memory of llm agents},
  author={Maharana, Adyasha and Lee, Dong-Ho and Tulyakov, Sergey and Bansal, Mohit and Barbieri, Francesco and Fang, Yuwei},
  booktitle={Proceedings of the 62nd Annual Meeting of the Association for Computational Linguistics (Volume 1: Long Papers)},
  pages={13851--13870},
  year={2024}
}

@inproceedings{personamem,
  title={Know Me, Respond to Me: Benchmarking LLMs for Dynamic User Profiling and Personalized Responses at Scale},
  author={Jiang, Bowen and Hao, Zhuoqun and Cho, Young Min and Li, Bryan and Yuan, Yuan and Chen, Sihao and Ungar, Lyle and Taylor, Camillo Jose and Roth, Dan},
  booktitle={Second Conference on Language Modeling}
}

@misc{qwen2025qwen25technicalreport,
      title={Qwen2.5 Technical Report}, 
      author={Qwen and : and An Yang and Baosong Yang and Beichen Zhang and Binyuan Hui and Bo Zheng and Bowen Yu and Chengyuan Li and Dayiheng Liu and Fei Huang and Haoran Wei and Huan Lin and Jian Yang and Jianhong Tu and Jianwei Zhang and Jianxin Yang and Jiaxi Yang and Jingren Zhou and Junyang Lin and Kai Dang and Keming Lu and Keqin Bao and Kexin Yang and Le Yu and Mei Li and Mingfeng Xue and Pei Zhang and Qin Zhu and Rui Men and Runji Lin and Tianhao Li and Tianyi Tang and Tingyu Xia and Xingzhang Ren and Xuancheng Ren and Yang Fan and Yang Su and Yichang Zhang and Yu Wan and Yuqiong Liu and Zeyu Cui and Zhenru Zhang and Zihan Qiu},
      year={2025},
      eprint={2412.15115},
      archivePrefix={arXiv},
      primaryClass={cs.CL},
      url={https://arxiv.org/abs/2412.15115}, 
}

@article{qwen3embedding,
  title={Qwen3 Embedding: Advancing Text Embedding and Reranking Through Foundation Models},
  author={Zhang, Yanzhao and Li, Mingxin and Long, Dingkun and Zhang, Xin and Lin, Huan and Yang, Baosong and Xie, Pengjun and Yang, An and Liu, Dayiheng and Lin, Junyang and Huang, Fei and Zhou, Jingren},
  journal={arXiv preprint arXiv:2506.05176},
  year={2025}
}

@inproceedings{haveliwala2002topic,
  title={Topic-sensitive pagerank},
  author={Haveliwala, Taher H},
  booktitle={Proceedings of the 11th international conference on World Wide Web},
  pages={517--526},
  year={2002}
}

@article{DBLP:journals/tois/ZhangDBMLCZDW25,
  author       = {Zeyu Zhang and
                  Quanyu Dai and
                  Xiaohe Bo and
                  Chen Ma and
                  Rui Li and
                  Xu Chen and
                  Jieming Zhu and
                  Zhenhua Dong and
                  Ji{-}Rong Wen},
  title        = {A Survey on the Memory Mechanism of Large Language Model-based Agents},
  journal      = {{ACM} Trans. Inf. Syst.},
  volume       = {43},
  number       = {6},
  pages        = {155:1--155:47},
  year         = {2025},
  url          = {https://doi.org/10.1145/3748302},
  doi          = {10.1145/3748302},
  bibsource    = {dblp computer science bibliography, https://dblp.org}
}

@article{DBLP:journals/corr/abs-2508-07407,
  author       = {Jinyuan Fang and
                  Yanwen Peng and
                  Xi Zhang and
                  Yingxu Wang and
                  Xinhao Yi and
                  Guibin Zhang and
                  Yi Xu and
                  Bin Wu and
                  Siwei Liu and
                  Zihao Li and
                  Zhaochun Ren and
                  Nikos Aletras and
                  Xi Wang and
                  Han Zhou and
                  Zaiqiao Meng},
  title        = {A Comprehensive Survey of Self-Evolving {AI} Agents: {A} New Paradigm
                  Bridging Foundation Models and Lifelong Agentic Systems},
  journal      = {CoRR},
  volume       = {abs/2508.07407},
  year         = {2025},
  url          = {https://doi.org/10.48550/arXiv.2508.07407},
  doi          = {10.48550/ARXIV.2508.07407},
  eprinttype    = {arXiv},
  eprint       = {2508.07407},
  bibsource    = {dblp computer science bibliography, https://dblp.org}
}

@article{Schlegelarticle,
author = {Schlegel, Katja and Sommer, Nils and Mortillaro, Marcello},
year = {2025},
month = {05},
pages = {},
title = {Large language models are proficient in solving and creating emotional intelligence tests},
volume = {3},
journal = {Communications Psychology},
doi = {10.1038/s44271-025-00258-x}
}

@inproceedings{DBLP:conf/nips/YangJWLYNP24,
  author       = {John Yang and
                  Carlos E. Jimenez and
                  Alexander Wettig and
                  Kilian Lieret and
                  Shunyu Yao and
                  Karthik Narasimhan and
                  Ofir Press},
  editor       = {Amir Globersons and
                  Lester Mackey and
                  Danielle Belgrave and
                  Angela Fan and
                  Ulrich Paquet and
                  Jakub M. Tomczak and
                  Cheng Zhang},
  title        = {SWE-agent: Agent-Computer Interfaces Enable Automated Software Engineering},
  booktitle    = {Advances in Neural Information Processing Systems 38: Annual Conference
                  on Neural Information Processing Systems 2024, NeurIPS 2024, Vancouver,
                  BC, Canada, December 10 - 15, 2024},
  year         = {2024},
  url          = {http://papers.nips.cc/paper\_files/paper/2024/hash/5a7c947568c1b1328ccc5230172e1e7c-Abstract-Conference.html},
  bibsource    = {dblp computer science bibliography, https://dblp.org}
}

@article{DBLP:journals/frai/PantiukhinSKJK25,
  author       = {Dmitrii Pantiukhin and
                  Boris Shapkin and
                  Ivan Kuznetsov and
                  Antonia Anna Jost and
                  Nikolay Koldunov},
  title        = {Accelerating earth science discovery via multi-agent {LLM} systems},
  journal      = {Frontiers Artif. Intell.},
  volume       = {8},
  year         = {2025},
  url          = {https://doi.org/10.3389/frai.2025.1674927},
  doi          = {10.3389/FRAI.2025.1674927},
  bibsource    = {dblp computer science bibliography, https://dblp.org}
}

@inproceedings{DBLP:BorgeaudMHCRM0L22,
  author       = {Sebastian Borgeaud and
                  Arthur Mensch and
                  Jordan Hoffmann and
                  Trevor Cai and
                  Eliza Rutherford and
                  Katie Millican and
                  George van den Driessche and
                  Jean{-}Baptiste Lespiau and
                  Bogdan Damoc and
                  Aidan Clark and
                  Diego de Las Casas and
                  Aurelia Guy and
                  Jacob Menick and
                  Roman Ring and
                  Tom Hennigan and
                  Saffron Huang and
                  Loren Maggiore and
                  Chris Jones and
                  Albin Cassirer and
                  Andy Brock and
                  Michela Paganini and
                  Geoffrey Irving and
                  Oriol Vinyals and
                  Simon Osindero and
                  Karen Simonyan and
                  Jack W. Rae and
                  Erich Elsen and
                  Laurent Sifre},
  editor       = {Kamalika Chaudhuri and
                  Stefanie Jegelka and
                  Le Song and
                  Csaba Szepesv{\'{a}}ri and
                  Gang Niu and
                  Sivan Sabato},
  title        = {Improving Language Models by Retrieving from Trillions of Tokens},
  booktitle    = {International Conference on Machine Learning, {ICML} 2022, 17-23 July
                  2022, Baltimore, Maryland, {USA}},
  series       = {Proceedings of Machine Learning Research},
  volume       = {162},
  pages        = {2206--2240},
  publisher    = {{PMLR}},
  year         = {2022},
  url          = {https://proceedings.mlr.press/v162/borgeaud22a.html},
  bibsource    = {dblp computer science bibliography, https://dblp.org}
}

@article{DBLP:journals/tois/HuangYMZFWCPFQL25,
  author       = {Lei Huang and
                  Weijiang Yu and
                  Weitao Ma and
                  Weihong Zhong and
                  Zhangyin Feng and
                  Haotian Wang and
                  Qianglong Chen and
                  Weihua Peng and
                  Xiaocheng Feng and
                  Bing Qin and
                  Ting Liu},
  title        = {A Survey on Hallucination in Large Language Models: Principles, Taxonomy,
                  Challenges, and Open Questions},
  journal      = {{ACM} Trans. Inf. Syst.},
  volume       = {43},
  number       = {2},
  pages        = {42:1--42:55},
  year         = {2025},
  url          = {https://doi.org/10.1145/3703155},
  doi          = {10.1145/3703155},
  bibsource    = {dblp computer science bibliography, https://dblp.org}
}

@inproceedings{DBLP:conf/ro-man/KasapM10,
  author       = {Zerrin Kasap and
                  Nadia Magnenat{-}Thalmann},
  editor       = {Carlo Alberto Avizzano and
                  Emanuele Ruffaldi},
  title        = {Towards episodic memory-based long-term affective interaction with
                  a human-like robot},
  booktitle    = {19th {IEEE} International Conference on Robot and Human Interactive
                  Communication, Viareggio, Italy, RO-MAN, 2010, September 13-15, 2010},
  pages        = {452--457},
  publisher    = {{IEEE}},
  year         = {2010},
  url          = {https://doi.org/10.1109/ROMAN.2010.5598644},
  doi          = {10.1109/ROMAN.2010.5598644},
  bibsource    = {dblp computer science bibliography, https://dblp.org}
}

@article{DBLP:journals/corr/abs-2502-06975,
  author       = {Mathis Pink and
                  Qinyuan Wu and
                  Vy Ai Vo and
                  Javier Turek and
                  Jianing Mu and
                  Alexander Huth and
                  Mariya Toneva},
  title        = {Position: Episodic Memory is the Missing Piece for Long-Term {LLM}
                  Agents},
  journal      = {CoRR},
  volume       = {abs/2502.06975},
  year         = {2025},
  url          = {https://doi.org/10.48550/arXiv.2502.06975},
  doi          = {10.48550/ARXIV.2502.06975},
  eprinttype    = {arXiv},
  eprint       = {2502.06975},
  bibsource    = {dblp computer science bibliography, https://dblp.org}
}

@article{articleMargaret2006,
author = {Svoboda, Eva and Mckinnon, Margaret and Levine, Brian},
year = {2006},
month = {02},
pages = {2189-208},
title = {The functional neuroanatomy of autobiographical memory: A meta-analysis},
volume = {44},
journal = {Neuropsychologia},
doi = {10.1016/j.neuropsychologia.2006.05.023}
}

@article{articleDickerson2009,
author = {Dickerson, Bradford and Eichenbaum, Howard},
year = {2009},
month = {09},
pages = {86-104},
title = {The Episodic Memory System: Neurocircuitry and Disorders},
volume = {35},
journal = {Neuropsychopharmacology : official publication of the American College of Neuropsychopharmacology},
doi = {10.1038/npp.2009.126}
}

@article{articleTanguay2023,
author = {Tanguay, Annick and Palombo, Daniela and Love, Brittany and Glikstein, Rafael and Davidson, Patrick and Renoult, Louis},
year = {2023},
month = {11},
pages = {},
title = {The shared and unique neural correlates of personal semantic, general semantic, and episodic memory},
volume = {12},
journal = {eLife},
doi = {10.7554/eLife.83645}
}

@inproceedings{Memoryos,
  title={Memory os of ai agent},
  author={Kang, Jiazheng and Ji, Mingming and Zhao, Zhe and Bai, Ting},
  booktitle={Proceedings of the 2025 Conference on Empirical Methods in Natural Language Processing},
  pages={25972--25981},
  year={2025}
}

\appendix

\section{Appendix}

\subsection{Experimental Settings}

Unless otherwise specified, all methods use Qwen2.5-72B-Instruct as the default LLM for memory construction and answer generation, and Qwen3-Embedding-8B as the embedding model. All LLMs in our experiments are accessed via commercial API services.

For ablation  and hyperparameter analysis experiments, we use Qwen3-30B-A3B to reduce computational cost. Since all HERO variants are evaluated under the same backbone and pipeline, these results reflect the relative contribution of each component rather than absolute scores comparable to the main results.

For all methods, answer generation and evaluation are conducted under identical prompts and LLMs. Other components follow the configurations described in their original papers. In particular, EverMemOS and Mem0 are evaluated using the EverMemOS evaluation pipeline, which provides a standardized and publicly available implementation for memory-based systems and ensures consistent preprocessing and metric computation.

For Mem0, we report the official hosted version in the main comparison, as it represents the strongest available implementation. 

On both the LoCoMo and PersonaMem benchmarks, the chunk size is set to 128. We set retrieval top-$k$ to 10 on LoCoMo and 40 on PersonaMem.

\begin{table}[t]
\centering
\small
\begin{tabular}{@{}llc@{}}
\toprule
\textbf{Backbone LLM} & \textbf{Method} & \textbf{ACC} \\
\midrule
\multirow{7}{*}{\textit{GPT-4o-mini}} 
  & MemoryOS$^\dagger$  & 0.547 \\
  & Mem0$^\dagger$      & 0.610 \\
  & MemU$^\dagger$      & 0.612 \\
  & MemOS$^\dagger$     & 0.759 \\
  & Zep$^\dagger$       & 0.811 \\
  & EverMemOS$^\dagger$ & 0.868 \\
  & \textbf{HERO}                & 0.870 \\
\cmidrule(lr){1-3}
\multirow{1}{*}{\textit{Qwen2.5-72B-Instruct}} 
  & \textbf{HERO}                & 0.880 \\
\cmidrule(lr){1-3}
\multirow{1}{*}{\textit{Qwen3-30B-A3B}} 
  & \textbf{HERO}                & 0.866 \\
\bottomrule
\end{tabular}
\caption{LoCoMo overall accuracy under different backbone LLMs. Results marked with $^\dagger$ are reported by EverMemOS~\cite{2026arXiv260102163H}; other results are from our evaluation.}
\label{tab:locomo_backbone_acc}
\end{table}

\begin{table}[t]
\centering
\small
\begin{tabular}{lcc}
\toprule
\textbf{Dataset} & \textbf{Top-$k$} & \textbf{ACC} \\
\midrule
\multirow{4}{*}{LoCoMo}
& 5   & 0.814 \\
& 10* & 0.828 \\
& 20  & 0.823 \\
& 40  & 0.818 \\
\midrule
\multirow{5}{*}{PERSONAMEM}
& 20  & 0.642 \\
& 30  & 0.659 \\
& 40* & 0.657 \\
& 50  & 0.667 \\
& 60  & 0.662 \\
\bottomrule
\end{tabular}
\caption{Sensitivity to retrieval top-$k$. Both datasets report accuracy.
Default values are marked with *.}
\label{tab:topk_sensitivity}
\end{table}

\begin{table}[t]
\centering
\small
\begin{tabular}{lccc}
\toprule
Dataset & Memory (MB) & Tr/Cu/Pr Nodes & Edges \\
\midrule
LoCo avg & 2.39 & 209/1163/1234 & 9181 \\
LoCo max  & 2.83 & 244/1398/1444 & 10941 \\
PM avg & 2.73 & 271/1645/1241 & 9810 \\
PM max & 3.35 & 338/2030/1528 & 12132 \\
\bottomrule
\end{tabular}
\caption{Per-user graph statistics of HERO.}
\label{tab:graph_stats}
\end{table}

\begin{table}[t]
\centering

\small
\begin{tabular}{lccc}
\toprule
Dataset & Total Time (s) & Cue Tok. & Profile Tok. \\
\midrule
LoCo & 294.5 & 73,278 & 108,991 \\
PM & 345.5 & 66,237 & 162,107 \\
\bottomrule
\end{tabular}
\caption{Graph construction overhead of HERO.}
\label{tab:construction_cost}
\end{table}

\subsection{Effect of Backbone LLMs}
Table~\ref{tab:locomo_backbone_acc} reports LoCoMo overall accuracy under
different backbone LLMs. Under the GPT-4o-mini setting, HERO achieves 0.870
accuracy, slightly higher than the strongest reported baseline EverMemOS
0.868. HERO also remains stable when using different Qwen backbones,
achieving 0.880 with Qwen2.5-72B-Instruct and 0.866 with Qwen3-30B-A3B.
These results suggest that HERO's gains are not tied to a single generation
model.

\begin{table}[t]
\centering
\small
\begin{tabular}{p{0.16\linewidth} p{0.78\linewidth}}
\toprule
\textbf{Stage} & \textbf{Content} \\
\midrule
\textbf{Query} &
``I've been thinking about ways to make my living space more unique and
personal. Any suggestions for incorporating interesting design elements?'' \\
\midrule
\textbf{Options} &
\textbf{(a)} You might consider incorporating natural elements... \newline
\textbf{(b)} For a personalized space, consider showcasing personal collections or heirlooms. These items not only add uniqueness but also reflect your personal history and tastes... \newline
\textbf{(c)} One intriguing approach could be exploring vintage or antique elements, such as old maps, blueprints, or historical artwork... \newline
\textbf{(d)} A dramatic approach could be using bold colors or geometric patterns. This could add a modern flair and energize your living area. Such elements often speak to personal style ... \\
\midrule
\textbf{Cue Activation} &
\textbf{Initial match:} \texttt{[LIVING\_SPACE, DESIGN\_ELEMENTS]} \newline
\textbf{Path A (Episodic Association):} \newline
Context: ``Assistant said, Collecting vintage travel maps enriches not only
your living space with...'' \newline
$\rightarrow$ Activated cues: \texttt{[VINTAGE\_TRAVEL\_MAPS]} \newline
$\rightarrow$ Associated cues: \texttt{[VINTAGE\_TRAVEL\_MAPS, VISUAL\_STORYTELLING]} \newline
\textbf{Path B (Profile Guidance):} \newline
Profile insight: ``User's hobby enriches their living space with visual
stories'' \newline
$\rightarrow$ Activated cues: \texttt{[HOBBY, VISUAL\_STORIES]} \newline
$\rightarrow$ Associated cues: \texttt{[HOBBY, VISUAL\_STORIES, VINTAGE\_MAPS]} \newline
... \\
\midrule
\textbf{Retrieved Episodic Traces} &
\textbf{T1 (User):} ``...I've started to appreciate the aesthetic value of collecting travel maps! They’re becoming beautiful decor in my home. The vibrant colors and intricate details found in each map really draw the eye, making every piece unique in its own right...'' \newline
\textbf{T2 (User):} ``I also started a collection of vintage travel maps to display in my home...'' \newline
\textbf{T3 (Assistant):} ``...Collecting vintage travel maps enriches not only your living space with visual stories but also your mind with historical context...'' \newline
... \\
\midrule
\textbf{HERO Answer} &
\textbf{Correct option:} ``One intriguing approach could be exploring vintage or antique elements, such as old maps, blueprints, or historical artwork. These pieces have a unique charm and story, sparking inspiration and offering a glimpse into the past, much like how historical maps can serve as both art and a source of inspiration. If you're interested, I can share more about how to source or incorporate such elements into your decor.'' \\
\bottomrule
\end{tabular}
\caption{Case study illustrating HERO's retrieval path on the PERSONAMEM benchmark, where episodic associations and profile guidance jointly activate relevant cues and retrieve supporting memories.}
\label{tab:case_dualpath}
\end{table}

\subsection{Selection of Retrieval Top-$k$}

We conduct a sensitivity analysis to examine whether HERO depends on a specific
retrieval top-$k$. As shown in Table~\ref{tab:topk_sensitivity}, HERO remains
stable across different top-$k$ values on both LoCoMo and PERSONAMEM. On
LoCoMo, top-$k=10$ achieves the best accuracy among the tested values. As for PERSONAMEM, it requires tracking dynamic profiles across multiple sessions, meaning evidence is highly scattered. 
The performance remains stable from top-$k=30$ to top-$k=60$; we use
top-$k=40$ as the default because it provides a practical coverage--cost
trade-off while retrieving fewer traces than larger settings.

\subsection{Scalability and Construction Cost}
\label{appendix:scalability}

We report the graph size and construction cost of HERO on LoCoMo and PERSONAMEM.

Table~\ref{tab:graph_stats} shows the average and maximum graph size per user. ``Tr'', ``Cu'', and ``Pr'' denote trace, cue, and profile nodes. The memory footprint stays below 3.4 MB per user in all cases.

Table~\ref{tab:construction_cost} reports the graph construction cost. On average, construction takes less than 6 minutes and fewer than 230K tokens per user. The graph is updated incrementally and does not require global recomputation.

\subsection{Prompt Templates}
\subsubsection{Profile Extraction}
\begin{agentbox}
### Role
You are a Knowledge Extraction System for a memory graph. Your goal is to extract key information from conversations to update the user's long-term memory.
Return JSON only: {"facts": ["...", "..."]} 
### What to Extract
Extract items into a flat list, covering these three categories:
1. **Facts**
One-time, specific actions or events.
Example: "User visited Tokyo"
2. **User Insights**
Situational states or emotions
Example: "User is currently injured", "User is currently feeling anxious about exams"
3. **human profiles**
Stable or recurring traits, habits, or preferences.
Example: "User prefers spicy food", "Tom is a software engineer"
### Rules
1. **Third-Person Perspective**: Convert all first-person dialogue into third-person objective facts.
2. **Stand-Alone Sentences**: The output must make sense **without context**. 
- BAD: "He liked *it*." 
- GOOD: "User liked *the sci-fi movie 'Dune'*."
- Replace pronouns (he, she, it, that) with specific entities.
3. **Atomic**: Each string must contain only ONE key piece of information.
4. **No Redundancy**: Do not extract "User said that..." or "User mentioned...". Just state the fact directly.
5. **Resolve Time**: If relative time is mentioned (yesterday, last week), try to make it specific contextually; otherwise, keep the original phrasing but ensure clarity.

### Example
Input:
Tom: "I'm exhausted. I stayed up until 3 AM coding that Python script."
AI: "Do you do this often?"
Tom: "Yeah, I'm a night owl. I usually work best after midnight."
Output:
{"facts": [
  "Tom stayed up until 3:00 AM coding a Python script",
  "Tom is currently exhausted",
  "Tom usually works best after midnight"]}

\end{agentbox}
\subsubsection{Answer Generation (LoCoMo)}
\begin{agentbox}
You are an intelligent memory assistant tasked with retrieving accurate information from episodic memories.
# DATA EXPLANATION:
The "Information" section contains original conversation fragments, each prefixed with a timestamp. 
Note that the information is NOT necessarily sorted chronologically. You must carefully distinguish the timestamps and speakers within each fragment.

# INSTRUCTIONS:
Your goal is to synthesize information from all relevant memories to provide a comprehensive and accurate answer.
You MUST follow a structured Chain-of-Thought process to ensure no details are missed.
Actively look for connections between people, places, and events to build a complete picture. Synthesize information from different memories to answer the user's question.

# CRITICAL REQUIREMENTS:
1. NEVER omit specific names - use "Amy's colleague Rob" not "a colleague"
2. ALWAYS include exact numbers, amounts, prices, percentages, dates, times
3. PRESERVE frequencies exactly - "every Tuesday and Thursday" not "twice a week"
4. MAINTAIN all proper nouns and entities as they appear
5. For time-related answers, use this format: 20 January 2026
6. Answer with exact words from the information context whenever possible
7. Double-check that your answer directly addresses the question asked
8. Ensure your final answer is specific and avoids vague time references

# Example:
Information: [2023-03-15]: Manry said, "I went to the vet yesterday."
Question: What day did Manry go to the vet?
Correct Answer: March 15, 2023
Explanation: Even though the phrase says "yesterday," the timestamp shows the event was recorded as happening on March 15th. Therefore, the actual vet visit happened on that date, regardless of the word "yesterday" in the text.

Information: [2023-03-15]: Manry said, "I went to the week before March 15, 2023."
Question: When did Manry go to the vet?
Correct Answer: the week before March 15, 2023.

## Output Format:
Please output the result in JSON format containing the following keys:
- **thought**: A step-by-step logical breakdown in the language of the input. Locate the specific sentence and identify the real names/dates.
- **answer**: The final, concise result (e.g., a specific name, date, or location) in the form of a short phrase. 

Example:
```json
{
"thought": "Scanning the memories...",
"answer": "20 January 2026"
}
```
\end{agentbox}
\subsubsection{Answer Generation (PERSONAMEM)}
\begin{agentbox}
# INSTRUCTIONS:
1. Carefully read all the conversation history provided as context
2. The question is a multiple-choice question with options (a), (b), (c), (d)
3. Select the option that **best demonstrates accurate memory of the user's history and preferences**.
4. **Prioritize specific, personalized responses over generic ones, but only if the historical details mentioned are factually correct based on the context.**
5. Pay attention to the reasons behind preference changes mentioned in conversations

# CRITICAL REQUIREMENTS:
1. Base your answer ONLY on the information provided in the context
2. If preferences have evolved, use the MOST RECENT preference
3. Verify that any "memories" mentioned in the options (e.g., "I remember you said...") strictly align with the provided text
4. Do NOT make assumptions beyond what is stated in the conversations

## Output Format:
- **Thought**: Analysis of relevant information from the conversation history
- **Answer**: Your final answer as a single letter: (a), (b), (c), or (d)
\end{agentbox}
\subsubsection{EVALUATION}
\begin{agentbox}
You are an expert grader that determines if answers to questions match a gold standard answer.
Your task is to label an answer to a question as 'CORRECT' or 'WRONG'. You will be given the following data:
(1) a question (posed by one user to another user),
(2) a 'gold' (ground truth) answer,
(3) a generated answer
which you will score as CORRECT/WRONG.

The generated answer may be more verbose. However, as long as the core information and factual claims are semantically consistent with the gold answer, it should be marked as CORRECT. Do not penalize for length or extra explanations.

Now it's time for the real question:
Question: {question}
Gold answer: {gold_answer}
Generated answer: {generated_answer}

First, provide a short (one sentence) explanation of your reasoning, then finish with CORRECT or WRONG.
Do NOT include both CORRECT and WRONG in your response. Just return the label CORRECT or WRONG in a json format with the key as "label".
\end{agentbox}

\subsection{Case study }
\label{sec:Case study}

As shown in Table~\ref{tab:case_dualpath}, this case illustrates how HERO’s dual-path retrieval integrates episodic evidence and profile guidance to support personalized reasoning. Starting from the initial cue LIVING SPACE, HERO activates two complementary paths. The episodic path surfaces concrete dialogue segments about vintage travel maps as home decor, while the profile path injects the user’s long-term preference for visual storytelling and hobby-driven decoration. These two paths converge on a shared semantic theme—personal collections as meaningful decor—which allows HERO to retrieve multiple supporting memories and correctly ground its decision on option (c). Option (a) suggests natural elements and (d) suggests bold colors or geometric patterns, neither of which is supported by the retrieved evidence. Option (b) refers to personal collections but remains too generic to capture the user’s specific interests. In contrast, option (c) directly aligns with the user’s core preference for vintage and historically grounded artifacts, and appropriately generalizes this preference to related design elements, making it the most well-supported choice.

\end{document}